\documentclass[letterpaper, 10 pt, conference]{ieeeconf}  % Comment this line out
\IEEEoverridecommandlockouts                              % This command is only
\usepackage{blindtext}
\usepackage{graphicx}
\usepackage{filecontents}
\usepackage{enumerate}
\usepackage{todonotes}

\usepackage[linesnumbered,algoruled,boxed,lined]{algorithm2e}
\SetKwInOut{Parameter}{parameter}
\usepackage{amsmath} % assumes amsmath package installed
\usepackage{amssymb}  % assumes amsmath package installed
\usepackage{hyperref}
\usepackage{array}
\title{\LARGE \bf
SwarmRob: A Toolkit for Reproducibility and Sharing of Experimental Artifacts in Robotics Research
}

\author{Aljoscha P\"{o}rtner$^{1,2}$, Martin Hoffmann$^{1}$ and Matthias K\"{o}nig$^{1}$% <-this % stops a space
\thanks{$^{1}$ The authors are with Campus Minden, Bielefeld University of Applied Sciences, 32427 Minden, Germany
        {\tt\small [forename].[surname]@fh-bielefeld.de}}%
\thanks{$^{2}$A. Poertner is with Faculty of Computer Science, Otto-von-Guericke-University, 39106 Magdeburg, Germany}
        \thanks{This work is financially supported by the German Federal Ministry of Education and Research (BMBF, Funding Number: 03FH006PX5)}% <-this % stops a space
}

\begin{document}

\maketitle
\thispagestyle{empty}
\pagestyle{empty}

%%%%%%%%%%%%%%%%%%%%%%%%%%%%%%%%%%%%%%%%%%%%%%%%%%%%%%%%%%%%%%%%%%%%%%%%%%%%%%%%
\begin{abstract}
Due to the complexity of robotics, the reproducibility of results and experiments is one of the fundamental problems in robotics research. While the problem has been identified by the community, the approaches that address the problem appropriately are limited. The toolkit proposed in this paper tries to deal with the problem of reproducibility and sharing of experimental artifacts in robotics research by a holistic approach based on operating-system-level virtualization. The experimental artifacts of an experiment are isolated in ``containers'' that can be distributed to other researchers. Based on this, this paper presents a novel experimental workflow to describe, execute and distribute experimental software-artifacts to heterogeneous robots dynamically. As a result, the proposed solution supports researchers in executing and reproducing experimental evaluations.

\end{abstract}

%%%%%%%%%%%%%%%%%%%%%%%%%%%%%%%%%%%%%%%%%%%%%%%%%%%%%%%%%%%%%%%%%%%%%%%%%%%%%%%%
\section{INTRODUCTION}
Because of the complexity of robotics, its research depends substantially on experimental evaluations to reveal consequences that cannot be seen from the start. Unfortunately, this complexity even makes the execution and reproduction of experiments very time-consuming and difficult. For example, the execution of an experiment with several robots, external sensors and actuators as well as additional monitoring systems needs the setup of various complex software artifacts. In many cases, these artifacts are exchanged by researchers across the globe, which makes it even more complicated to execute and, above all, reproduce an experiment. Some research networks have established special interest groups, e.g. EURON Good Experimental Methodology (GEM) or the IEEE Technical Committee on Performance Evaluation and Benchmarking of Robotic and Automation Systems (TC-PEBRAS), to develop methodologies and tools that improve the situation. One outcome is the demand by Bonsignorio and del Pobil~\cite{bonsignorio2015}, who call for a possibility to practically replicate results for validation. The authors suggest a new community-wide agreement on the content of a research paper in robotics. They propose the additional accompaniment of \textit{data sets}, \textit{code identifiers} and \textit{hardware identifiers} to provide a level of transparency where the experiment itself can be replicated in another environment by an objective researcher. Guglielmelli~\cite{guglielmelli2015} points out that reproducibility is one of the critical factors for dependable robots and supports the trust of the society in such systems. To increase the maturity of robotics research, the assurance of reproducibility and the possibility of objective disconfirmation is indispensable. This paper proposes a novel workflow that addresses the requirements stated by Bonsignorio and del Pobil~\cite{bonsignorio2015} and introduces a framework that abstracts the complexity of experimental setups by using containerization technologies. The proposed approach makes it significantly easier to share the experimental artifacts and reproduce the environment of the inital experiment.

\begin{figure}
\centering
\includegraphics[width=0.35\textwidth]{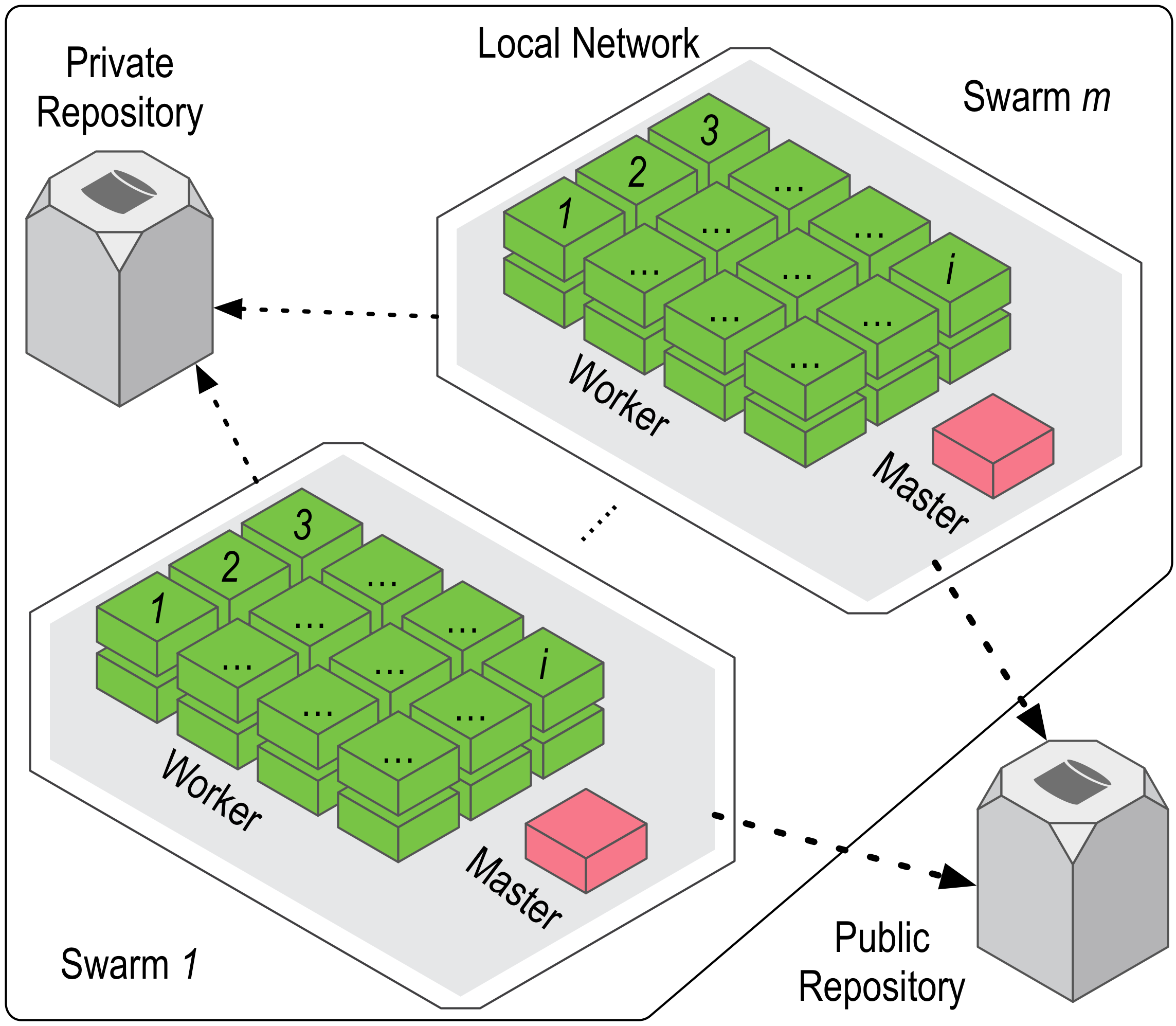}
\caption{The Architecture of \textit{SwarmRob} - The green cubes represent the worker nodes and the red cubes represent the master nodes. Every bounding box illustrates a swarm.The outer box illustrates the local network of the laboratory and the grey boxes illustrates the repositories where the worker can obtain the definition files.}
\label{fig:sr_architecture}
\end{figure}

\section{RELATED WORK}
The following section discusses the related work that is focused on systems consisting of multiple robots, namely \textit{Multi-Robot Testbeds} and \textit{Application Containerization}.
\subsection{Multi-Robot Testbeds}
 \par In some branches of robotics, e.g. evolutionary robotics, it is common to evaluate approaches by simulating real-world experiments. Unfortunately, this approach leads oftentimes to differing results, compared to results of the system in the real world~\cite{koos2013}. To synergize the evaluation capabilities of simulation evironments and equally prevent the ``reality gap``, testbeds for multiple robots have been developed. Michael et al.~\cite{michael2008} introduced a testbed for large multi-robot systems with a strong focus on classical swarm robotics and homogeneous compositions. The system is limited to two different robot platforms and supports the integration of a three-dimensional simulation environment. Johnson et al.~\cite{Johnson2006} proposed one of the first remotely accessible testbeds for mobile robots combined with sensor networks. The testbed uses a client-server architecture with a web-based front-end for creating and managing experiments, including logging and remote code execution. Additionally, it supports robot localization using external cameras and remote robot control. A very similar approach called \textit{HoTDeC} were introduced by Stubbs et al.~\cite{Stubbs2006}. While \textit{HoTDeC} also uses external cameras, it is designed for homogeneous composition of autonomous hovercrafts. Another variant, which is designed for cheaper and smaller robots, were proposed by Pickem et al.~\cite{Pickem2017}. The system relies on very small robots called \textit{GRITSbots}~\cite{Pickem2015} and should also be available to external users for remotely carrying out their experiments. The disadvantages of testbeds are that they are limited to a homogeneous set of components and are usually inflexible to technical expansions. Additionally, the possibility to reproduce experiments is limited to a very small subset of researchers which are in possession of the complete testbed. A more inclusive and flexible alternative referred to in the literature is the use of containerization technologies.
%Some more examples

\subsection{Containerization in Multi-Robot Systems}
 A more integral approach to the problem of reproducibility is the use of container technologies to encapsulate the complexity of software and hardware. Boettiger~\cite{boettiger2015} and Cito et al.~\cite{cito2016} discussed the relevance of containers for software engineering research and mention four technical challenges that prevent reproducibility in software engineering reserach that are also problems in robotics research: (\textit{i.}) the \textit{``Dependency Hell''}, which is the problem of reproducing computational environments to run the software, (\textit{ii.}) \textit{Imprecise Documentations}, which multiplies the problem \textit{(i.)} and is therefore another barrier to install and run the software, (\textit{iii.}) \textit{Code erosion}, which is the problem of running outdated or updated code in current environments and (\textit{iiii.}) \textit{Barriers to Adoption with Existing Solutions}, which is the problem that existing technological solutions that would solve some of the problems need a high level of expertise and are therefore neglected by the researchers. Both propose containers as an approach to face these challenges. Currently, the use of containers in robotics research is not very popular because a practical, generic approach is missing. One of the first approaches that addressed this deficiency and proposed an experimental workflow in robotics research inspired by containerization is the \textit{Cognitive Interaction Toolkit (CITk)} by Lier et al.~\cite{lier2014}. The approach uses containerization in an automated build and deployment process for simulation environments. An extension to the \textit{CITk} is the \textit{RoboBench} project presented by Weisz et al.~\cite{weisz2016} that extends the \textit{CITk} by a benchmarking suite that allows the distribution of system-wide benchmarking containers via public repositories. Unfortunately, the initial approach as well as the extension do not support physical robots.

\begin{figure*}[t]
\centering
\includegraphics[width=1\textwidth]{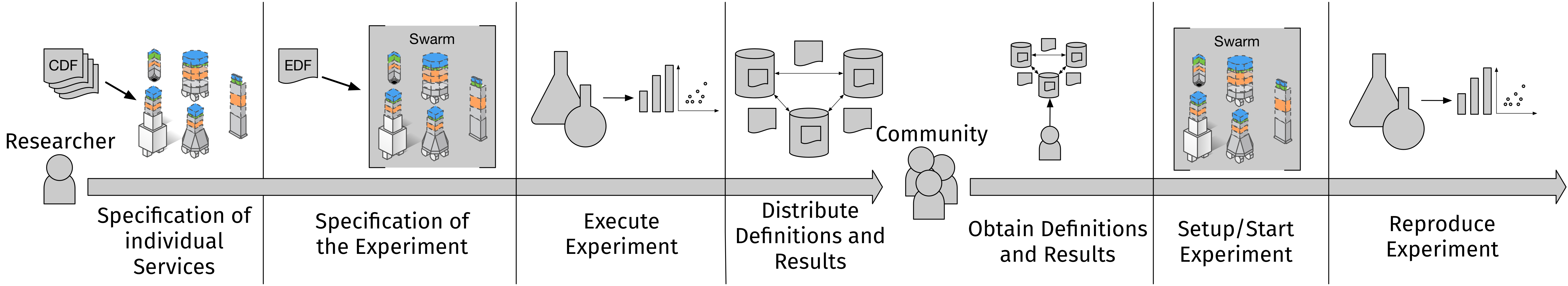}
\caption{Workflow of SwarmRob - The figure illustrates the research phase (left timeline) and the review phase (right timeline) of the workflow with their related subphases.}
\label{fig:workflow}
\end{figure*}

%TODO Mehr Gegenüberstellung zu den anderen Lösungen
\section{CONTRIBUTION TO THE STATE OF THE ART}
In order to address the previously discussed deficiency, this paper presents a novel framework that enables the distribution of experimental artifacts in robotics research based on a container technology, called \textit{Docker}~\cite{docker2017}. It supports the dynamic assignment of containers to nodes based on predefined composition definitions, which can be easily provided to other researchers. Unlike other orchestration solutions, the system is tailored to robotics and incorporates the underlying hardware as a parameter within the service allocation process. In the previous section, some of the technical challenges that prevent reproducibility have been discussed. Below, the particular challenges are revisited and solution concepts of the approach will be discussed:
\begin{enumerate}[(i)]
\item \textbf{Imprecise Documentations}: The approach tackles the problem of \textit{Imprecise Documentations} by detailed definition files. They are the important basis for the bootstrapping of the experiment.

\item \textbf{``Dependency Hell``}: Because of the packaging of software and dependencies in containers, the approach addresses the several forms of the \textit{Dependency Hell} appropriately .

\item \textbf{Code erosion}: The \textit{Code erosion} in robotics applications is addressed by the use of software containers, which make it possible to use outdated software via virtualized operating system environments.

\item \textbf{Barriers to Adopting Existing Solutions}: Because the use of container technology in robotics needs a high level of expertise and therefore raises \textit{Barriers to Adopt the Solution}, the system automates most of the necessary steps, e.g. container distribution, virtualized networking, and thus reduces obstacles to use the technology.

\end{enumerate}

The previous section has highlighted the contribution to the state of the art as well as the advantages of the proposed solution comparing to other approaches. The following section describes the system in detail.

\section{SWARMROB}
The \textit{SwarmRob} system simplifies the re-execution and reproducibility of experiments in robotics research. Therefore, it uses containers for the deployment of robotics applications and bootstrap as well as orchestration mechanisms to abstract the complexity of the technology. The remainder of this section discusses the several partial aspects in detail.

\subsection{Definitions}
For clarification, several notions need to be defined: An experiment is a composition of services $z \in Z$, where $Z$ is finite set of services. A service $z$ is a detailed description of the assembly of a specific artifact of the experiment. A \textit{swarm} is an element $s \in S$, where $S$ is a finite set of swarms and $s$ itself is also a finite set which means that $S$ is a set of sets. The atomic elements of $s$ are agents $a \in A$, which implies that $s \subset A$.  Every agent in a swarm has either the role of a \textit{worker} $w \in W$, where $W \subset A$ or the role of a \textit{master} $m \in M$, where $M \subset A$ with the restriction that $M$ is a singleton set. As a consequence, one has $A \setminus (M\cup W) = \emptyset$. Furthermore, the amount of workers is the amount of robots in the swarm less the \textit{master} $W \approx (A\setminus M)$ and no participant can be worker and master at the same time $(M\cap W) = \emptyset$.

\subsection{The SwarmRob Workflow}
Along with the framework, \textit{SwarmRob} proposes a novel workflow for the documentation, distribution and execution of experiments (Fig. \ref{fig:workflow}). The workflow of the system can be subdivided in two meta-phases: the \textit{(i.) Research Phase}, where the initial experiment is constructed and evaluated and the \textit{(ii.) Review Phase}, where the results are reproduced and reviewed by the community. The research phase is subdivided in four consecutive phases: The first phase is concerned with the \textit{Specification of Individual Services}, where the hardware and software configurations of services are documented via \textit{CDFs (Container Definition Files)}. The \textit{CDF} can be interpreted as a basic description of an artifact of the experiment. The \textit{CDF} describes e.g. the operating system of the service, the required software packages or the code repositories that should be cloned within the initialization. The deliverable of the specification phase is a service that is ready for execution. A service can be reused in another experiment, by forking its \textit{CDF}. The second phase of the workflow is the \textit{Specification of the Experiment}. In this phase, the composition of the services, which is the basis of the experiment, is defined. The specification is deposited in a so called \textit{EDF (Experiment Definition File)}. The \textit{EDF} refers to the \textit{CDFs} and adds additional information like network configurations, relationships between services (e.g. two services should be executed on the same machine) and hardware requirements. The deliverable of this phase is a complete specification of an experiment composition that can be re-executed in various environments. The third phase of the workflow is the actual experiment. The experiment is performed by the help of \textit{SwarmRob} but is completely independent of the framework in terms of evaluation and monitoring. The deliverable of this phase is the result of the experiment that is planned to be shared, e.g. in a scientific publication. In the subsequent phase, the researcher can make the composition of the experiment (CDFs and EDF) and the results of the experiment available to other researchers using public repositories and academic publications. In the \textit{Review Phase}, the community can easily obtain the results and definitions and is able to review the performed experimental evaluations. \textit{SwarmRob} automates the distribution of the containers, the initialization of the robots as well as the configuration of the inter-robot network. The deliverable of this phase is a bootstrapped system that resembles the situation of the inital experiment as close as possible. Based on this, the experiment can be reproduced and a much more qualified feedback for the author is possible.

\subsection{The Container Ecosystem}
\begin{figure}
\centering
\includegraphics[width=0.35\textwidth]{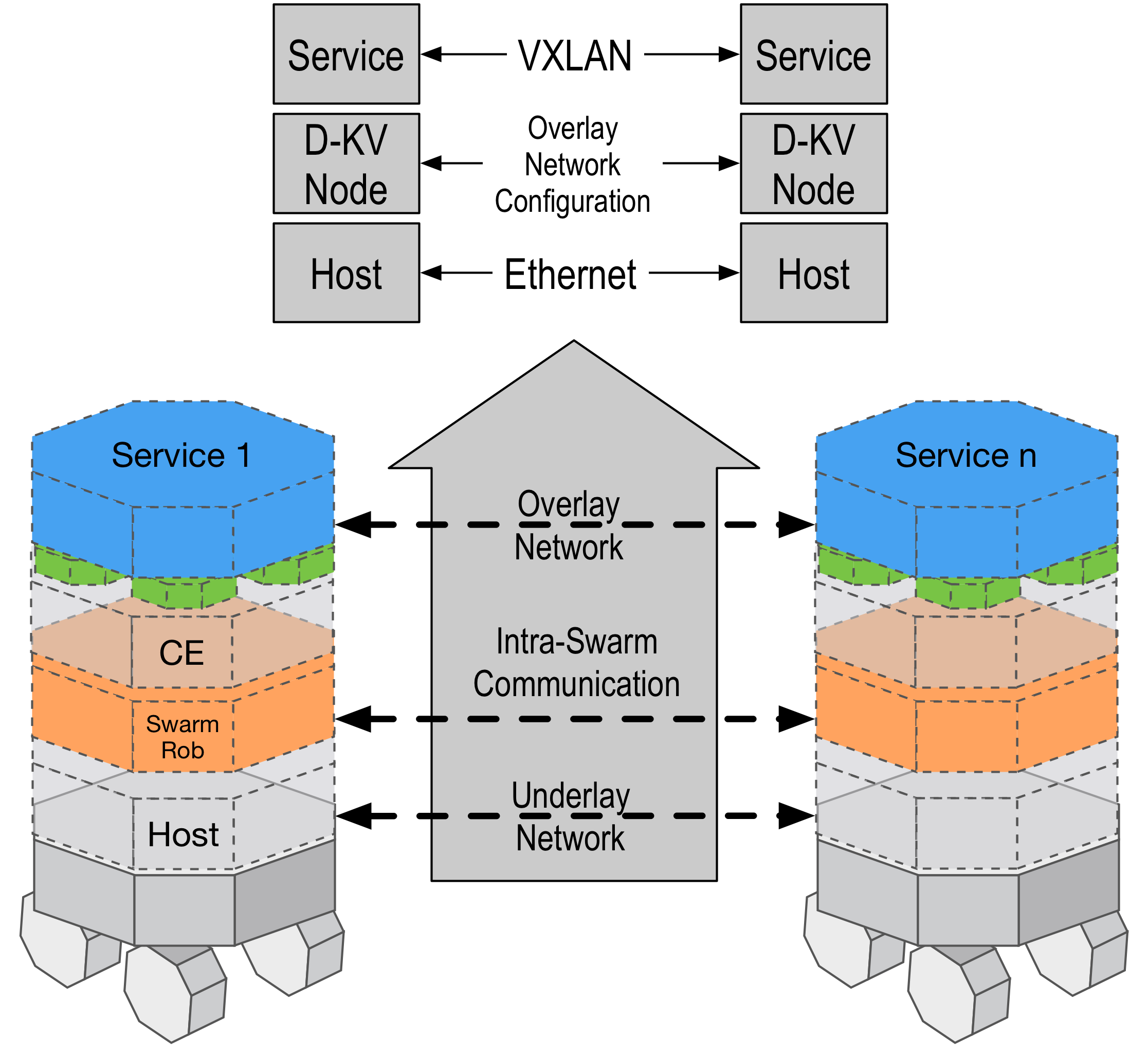}
\caption{System and Inter-Robot Network Architecture using Overlay Networks - The \textit{Underlay Network} represents the physical network connection between the hosts, the \textit{Intra-Swarm Communication} represents the commands and information exchanged between the participants of a swarm and the \textit{Overlay Network} is the communication channel used for the communication between containerized applications.}
\label{fig:network_architecture}
\end{figure}

The packaging of software using containers is founded on the idea of abstracting and isolating hardware and encapsulating software. It guarantees that any process inside a container cannot see a process outside of the container. Every container can be easily distributed and executed using a \textit{CDF} that defines among other things, the operating system, required software packages and dependencies as well as mounted volumes. The behaviour of each container is managed by a \textit{container engine (CE)} that handles its life cycle. In \textit{SwarmRob}, the container technology is used to encapsulate the services running on each robot. Using the example of the \textit{Robot Operating System (ROS)}, a high amount of software dependencies and configuration is required to replicate a system. The container technology of \textit{SwarmRob} simplifies this since it manages all of the dependencies and configurations in one single \textit{CDF} that can be easily distributed and executed on every robot with the same hardware configuration. Because the manual setup of large-scale containerized systems is time-consuming and needs expertise, \textit{SwarmRob} takes care of the necessary steps. For this purpose, every node of the system runs a \textit{SwarmRob} daemon. The daemon automates all aspects of the container workflow, e.g. obtain and start containers, setup networks or provide system information of the worker to the master. Because of the highly distributed architecture of system, the daemon also manages the distributed access of information used for inter-service networking and controls the service allocation process participation. The technology stack is illustrated by the left robot shown in Fig. \ref{fig:network_architecture}.
%\begin{figure}
%\centering
%\includegraphics[width=0.35\textwidth]{figures/container_ecosystem}
%\caption{Software and hardware architecture of a single robot within \textit{SwarmRob} - The Figure should illustrate the technology stack, the isolation of containers and its easy distribution to other robots or components}
%\label{fig:container_ecosystem}
%\end{figure}
\subsection{Network Architecture}
Information exchange between robots is one of the originating aspects of multi-robot systems. \textit{SwarmRob} supports the communication between services using \textit{VXLAN}~\cite{rfc7348}. It encapsulates the actual communication via a tunneling mechanism on top of the actual physical infrastructure. \textit{VXLAN} uses the existing underlay network and encapsulates the traffic in UDP packets of the underlay network and adds an additional \textit{VXLAN} header. In a \textit{VXLAN}-based overlay network, only the \textit{VTEPs} (\textit{VXLAN Tunnel Endpoints}) can encapsulate, respectively decapsulate the traffic. In the context of \textit{SwarmRob}, the \textit{VTEPs} of an overlay network are the services that assemble an experiment. For that purpose, additional information, e.g. network configuration and the members of the communication, need to be shared between the \textit{VTEPs}. In the toolkit, this is implemented using a \textit{distributed key-value store (D-KV)} that guarantees the conflict-free and highly topical access to this information. The isolated experiment-specific network traffic has several benefits: First of all, due to the isolation, the monitoring of the communication traffic is much simpler, because potential noise (e.g. broadcast, web traffic) of the host system is filtered. This results in a better reproducibility of communication experiments and comparability of metrics. In addition, the risk of potential network conflicts (e.g. already used resp. blocked ports or IPs, misconfigured networks or communication channels) can be reduced, because the network traffic as well as the configuration (IPs, Ports, Subnets) of the overlay network tunnel is completely independent of the underlay network and unique for each instance of the overlay network. The communication between the master and its workers within a swarm is implemented using remote procedure calls. The network architecture of the system is illustrated in Fig.~\ref{fig:network_architecture}.

\subsection{The SwarmRob Architecture}
The architecture of \textit{SwarmRob} considers the special requirements of multi-robot experiments by using a master-worker architecture, where every participant in the system is either a worker that can run one or more services or a master that controls the swarm. Every swarm of the system is a unique instance initialized by its master node. The master node is an indespensable component of the swarm. It manages the lifecycle of the workers, the service allocation process as well as the overall logging of the system. Every instance of a swarm can be joined as a worker node by its unique identifier and the network address of the master. If the swarm should execute an experiment, the master loads the \textit{EDF} and allocates the services dynamically to the workers based on their hardware capabilities as well as their workload. The worker that is used to run a specific service obtains the \textit{CDF} of the assigned service either by a publically or locally accessible repository. The whole architecture of the system is illustrated in Fig.~\ref{fig:sr_architecture}.
\begin{figure}
\centering
\includegraphics[width=0.42\textwidth]{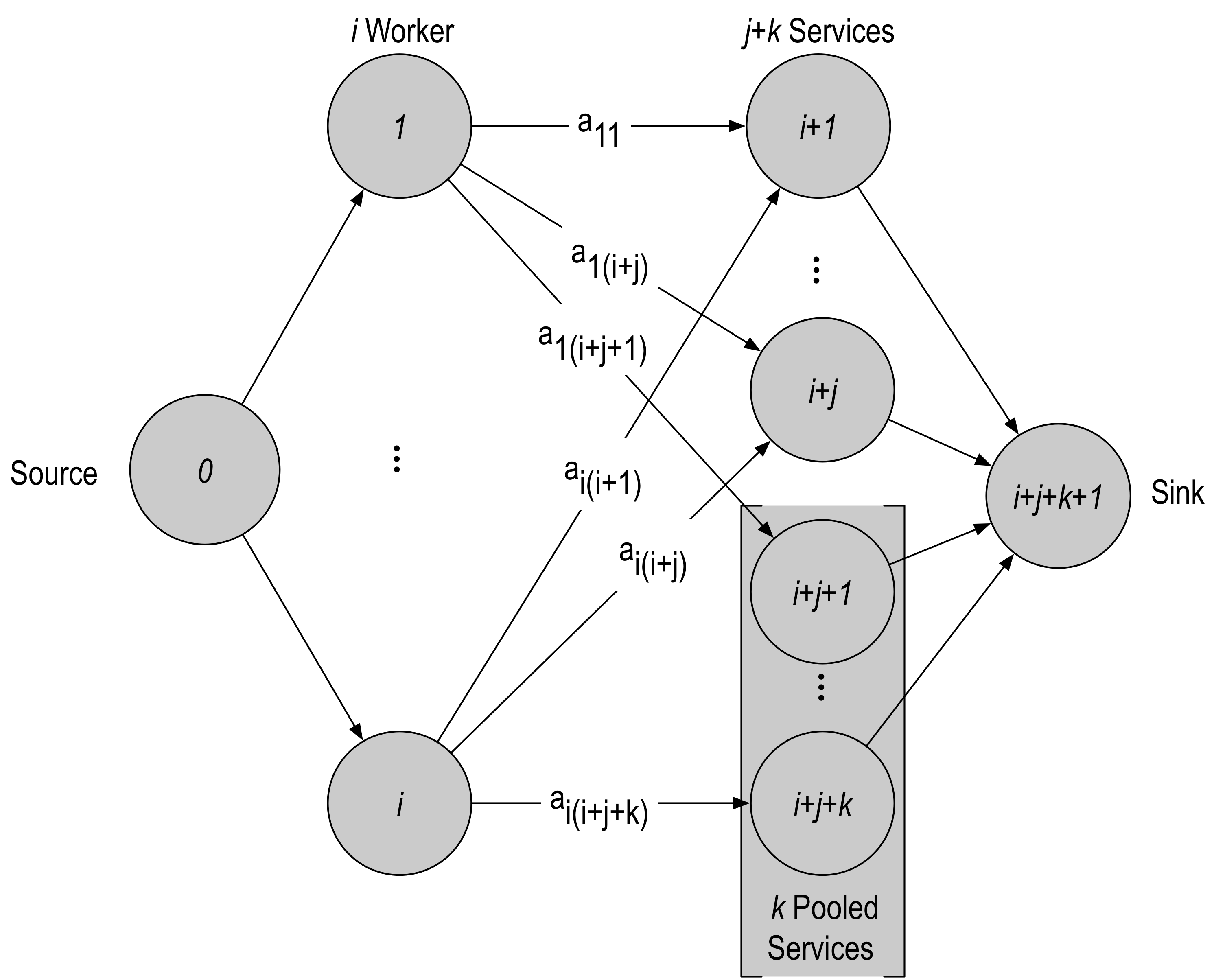}
\caption{Model of the Service Allocation Problem as a Flow Network - The nodes on the left side represent the set of worker, the node on the right side represent the set of services and pooled services. The edge labels represent the costs to run a service \textit{j+k} on a worker \textit{i}.}
\label{fig:container_allocation_graph}
\end{figure}
\subsection{Dynamic Service Allocation}
\begin{figure}[!t]
 \centering
 \includegraphics[width=0.49\textwidth]{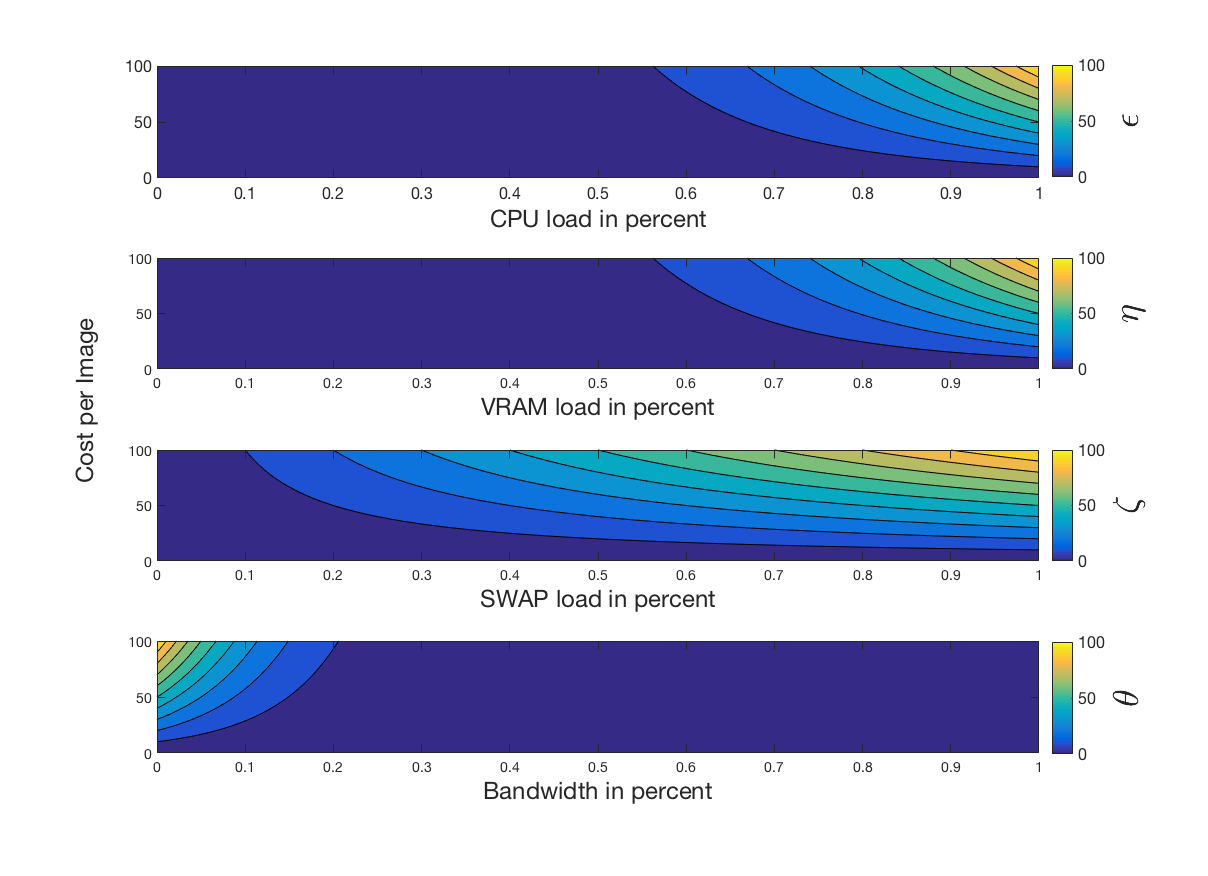}
 \caption{Contour Plots of the Workload Cost Functions - The several x-axes represent the relative workload, the y-axes represent the pre-defined costs per image and the color of the contour plot indicates the resulting cost.}
 \label{fig:contour_plots}
\end{figure}

The experiment bootstrap process is implemented using a dynamic allocation of services to workers. The problem is an instance of the \textit{assignment problem} and can be modelled as a flow network (Fig.~\ref{fig:container_allocation_graph}) and formulated as a maximum-flow with minimum-cost problem. A flow network is a directed graph $G=(V,E)$ with a source vertex $s \in V$ and a sink vertex $t \in V$. Each edge has a capacity $c:E \rightarrow \mathbb{R}^{+}$, denoted by $c_{uv}$ that represents the maximum amount of flow that can pass through an edge. The costs $a:E \rightarrow \mathbb{R}^{+}$ to run a container are denoted by $a_{uv}$. A flow is a mapping $f:E \rightarrow \mathbb{R}^{+}$, denoted by $f_{uv}$ subjected to a capacity constraint (Equation~\ref{eq:capacity_constraint}) that guarantees that an arc's flow cannot exceeds its capacity and a conservation constraint that equals Kirchhoffs current law (Equation~\ref{eq:conservation_constraint}).
\begin{subequations}
\begin{alignat}{2}
&\forall (u,v) \in E: f_{uv} \leq c_{uv}\label{eq:capacity_constraint}\\
&\forall  v\in V \setminus \{s,t\}: \sum_{u:(u,v)\in E}{f_{uv}} =\sum_{u:(v,u)\in E}{f_{vu}}. \label{eq:conservation_constraint}
\end{alignat}
\end{subequations}
The value of the flow (amount of flow passing from source to sink) is defined by
\begin{flalign}
|f|=\sum_{v:(s,v)\in E}{f_{sv}} - \sum_{v:(v,s)\in E}{f_{vs}}\text{.}
\end{flalign}
The total cost over all edges is given by
\begin{flalign}
|a| = \sum_{(u,v) \in E}{a_{uv} \cdot f_{uv}}\text{.}
\end{flalign}
The maximum-flow with minimum-cost problem is to maximize $|f|$ while minimizing $|a|$.
Each worker $w \in W$ and each service $z \in Z$ is a vertex $\omega : (W\cup Z) \rightarrow V$ subjected to the constraint that an edge is only allowed between a worker and a service.
%\begin{equation}
%\forall (u,v) \in E: \{(u,v) | u \in W, v \in C\}
%\end{equation}
Whether two vertices are connected by an edge is given by the hardware capabilities of that worker, that is, if the worker is able to run the specified service. This restriction on the edges is specified via a binary matrix $H$, where $R$ is a binary relation $R \subseteq W\times Z$ that holds iff. a worker $w_i \in W$ has the hardware capabilties to run a service $z_j \in Z$. As a consequence, the entries of $H$ are defined by
\begin{equation}
H_{i,j}=\begin{cases}
1 \text{~~}(w_i,z_j) \in R \\
0 \text{~~}(w_i,z_j) \notin R\text{.}
\end{cases}
\end{equation}
The corresponding transformation function is a mapping $\kappa: Z \times W \times H \rightarrow E$. The overall costs $a_{uv}$ for running a service on a worker is given by the workload of the worker and the pre-defined costs for a single container. The overall workload of the worker is composed of the CPU load denoted by $\epsilon$, the VRAM load denoted by $\eta$, the SWAP load denoted by $\zeta$ and the bandwith of the worker denoted by $\theta$, all multiplied by a weighting factor $\delta$:
\begin{flalign}
a_{uv} = \epsilon_{uv} \cdot \delta_{\epsilon} + \eta_{uv} \cdot \delta_{\eta} + \zeta_{uv} \cdot \delta_{\zeta} +\theta_{uv} \cdot \delta_{\theta},
\end{flalign}
such that the sum of all weights $\delta$ equal 1. The single cost of each load is given by their related cost functions:
\begin{subequations}
\begin{flalign}
\epsilon_{uv} &= \alpha \cdot \beta^4 \textit{~(CPU load)}\\
\eta_{uv} &= \alpha \cdot \beta^4 \textit{~(VRAM load)}\\
\zeta_{uv} &= \alpha \cdot \beta \textit{~(SWAP load)}\\
\theta_{uv} &= \alpha \cdot (1-\beta)^{4} \textit{~(Bandwidth),}
\end{flalign}
\end{subequations}

where  $\alpha \in \{x \in \mathbb{R}|0\leq x \leq 100\}$ is the predefined costs for the service and $\beta \in \{x \in \mathbb{R}|0\leq x \leq 1\}$ is the currently measured relative workload. Fig.~\ref{fig:contour_plots} illustrates the cost functions as contour plots. It shows that $\epsilon$ and $\eta$ are progressive cost functions, where the exponent is chosen such that the cost increase with a strong degree of progressivity in the second half of the workload scale to relieve high-loaded workers. Because of the assumption that a high swap indicates a very high workload of some worker, the exponent of the cost function $\zeta$ is chosen such that the cost rise earlier than $\epsilon$ and $\eta$. In the case of $\theta$, the cost function is chosen such that the cost decrease along with the available bandwith of the worker. If the bandwith tends to zero, the cost increases. The cost calculation algorithm is designed as a distributed algorithm that allows uses parallelization of the cost calculation. We assume that $m$ is the total number of agents, $n$ is the total number of services, $i$ lies in the interval $[1,m]$ and $j$ lies in the interval $[1,n]$. The cost per worker and per service are represented as a cost matrix $A$ of size $m \times n$, where the entries of $A$ are defined by $A_{i,j}=a_{uv}$. Each worker can run exactly one container, which implies that each capacity is given by $c_{uv} = 1$. The overall capacities are represented as a matrix $C$ of size $m \times n$, where the entries of $C$ are defined by $C_{i,j}=c_{uv}$.

In order to enable that two or more services can be executed on one worker, the system supports \textit{pooled services} $\Phi$, where $ \Phi \subset \mathcal{P}(Z)$. To take into account that in some cases (e.g. when all machines have a high workload) the execution of a pooled service on different machines is more favorable than the execution on a single machine, the set of service vertices is the set of the single services united with the set of the pooled services such that $\tau: (W\cup Z \cup \Phi) \rightarrow V$. We assume that $k$ lies in the interval $[1,n]$. The dependencies of services are represented via a binary matrix $\Xi$ of size $n \times n$, where $D$ is a binary relation $D \subseteq Z \times Z$ that holds iff. a service $z_k \in Z$ depends on a another service $z_j \in Z$.  The entries of $\Xi$ are given by
\begin{equation}
\Xi_{k,j}=\begin{cases}
1 \text{~~}(z_k,z_j) \in D \\
0 \text{~~}(z_k,z_j) \notin D\text{.}
\end{cases}
\end{equation}
The cost of a pooled service $\phi \in \Phi$ for a worker $w \in W$ is given by the sum of cost of their respective containers, multiplied by a discount factor $\delta$.
%\begin{subequations}
%\begin{flalign}
%a_{u\phi} = \delta \cdot \sum_{(u,v) \in E}{a_{uv}}\text{,~}v %\in \phi, \delta \in \mathbb{R}^{+}, \delta \leq 1
 %\end{flalign}
%\end{subequations}
The corresponding cell of the hardware capability matrix $H$ of a pooled container $\phi \in \Phi$ for a worker $w \in W$ is the logical conjuction of all entries of $H$ for that service.
%\begin{equation}
%\land: W \times \Phi \rightarrow \{0,1\}, (w,\phi) \mapsto %\land(w,\phi)
%\end{equation}
The entries of the capacity matrix $C$ stays the same as for single services.
The problem is solved using the \textit{cost-scaling push-relabel (CSPR)} algorithm by Goldberg and Kennedy~\cite{goldberg1997} implemented in the Google Optimization Toolbox~\cite{google2017}.
%\begin{equation}
%\bordermatrix{
 % & 1	& \dots   & n   \cr
%1 & h_{11} & \dots & h_{n1} \cr
%\vdots & \vdots & \ddots & \vdots \cr
%m & h_{1m} & \dots & h_{nm} \cr
%m + 1 & h_{1(m+1)} & \dots & h_{n(m+1)} \cr
%\vdots & \vdots & \ddots & \vdots \cr
%m + j & h_{1(m+j)} & \dots & h_{n(m+j)} \cr
%}
%\label{eq:hw_matrix}
%\end{equation}

\begin{figure}[b]
 \centering
 \includegraphics[width=0.35\textwidth]{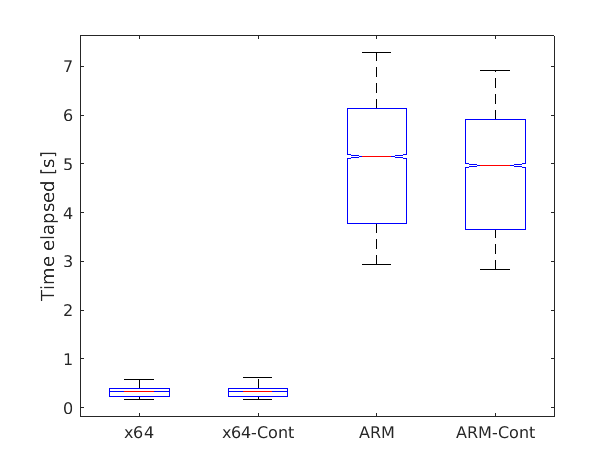}
 \caption{Boxplots of the Elapsed Time in Seconds of Container and Native \textit{SLAMBench}, each for x64 and ARM.}
 \label{fig:slambench}
\end{figure}

\section{EVALUATION}
The following section discusses the results of the evaluation. To reveal possible differences between system architectures, the experiments were performed using an x64-based system consisting of an \textit{Intel Core i5-6200U} with \textit{2.30 GHz}, \textit{4096 MB LPDDR3} and \textit{IEEE 802.11ac-WLAN} connection and an ARM-based system consisting of an \textit{ARM Cortex-A7} with \textit{900 MHz}, \textit{1024MB DDR2-SDRAM} and a \textit{IEEE 802.3-10/100-Mbit/s} connection. The evaluation of the service allocation algorithm as well as the overall system evaluation were performed using the previously described ARM-based system composed as a 12-node cluster.

\begin{figure*}[t]
\centering
\includegraphics[width=1\textwidth]{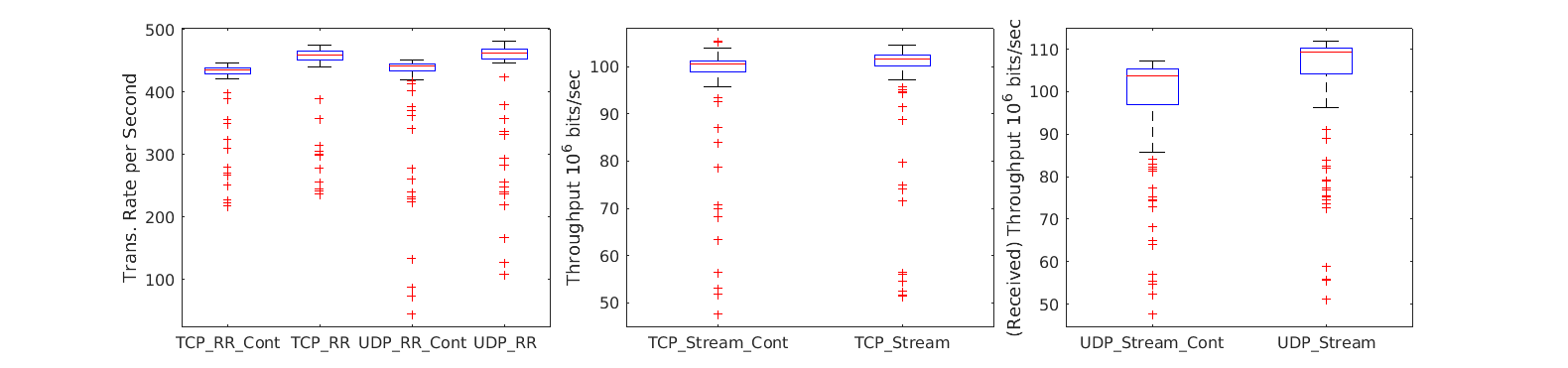}
\caption{Boxplots of Network Performance Measures - The left figure shows the TCP/UDP Request/Response Performance, the figure in the middle shows the TCP Stream Performance and the right figure shows the UDP Stream Performance, each for the containerized and the non-containerized version}
\label{fig:netperf}
\end{figure*}

 \begin{table}
 \setlength\extrarowheight{2.5pt}
    \caption{Statistical Dispersion of the Container Performance Experiment}
    \label{tab:msd_container_performance}
    \centering
    \begin{tabular}{|p{4cm}|l|l|}
    \hline
    \textit{Metric} & \textit{Std. Deviation} & \textit{CV}\\ \hline \hline
    \textit{x64} & 0.0908 & 0.2800  \\
    \textit{x64\_Cont} & 0.0908 & 0.2800 \\
    \textit{ARM} & 1.2258 & 0.2429 \\
    \textit{ARM\_Cont} & 1.1820 & 0.2426 \\ \hline
    \end{tabular}
   \end{table}
\subsection{Container Performance}
An important aspect is the question whether the containerization of robotics applications affect the computational performance of the system. To consider the special requirements of robotics applications and to ensure reproducibility, the evaluation was performed using \textit{SLAMBench}~\cite{nardi2015}, which is a framework for quantifying \textit{SLAM} algorithms. Additionally, the framework contains implementations of \textit{SLAM} algorithms and it is able to run on pre-recorded datasets. For the performance evaluation of the containerized case compared to the native case, the experiment was performed using the \textit{KinectFusion} algorithm~\cite{newcombe2011} executed on the \textit{ICL-NUIM} dataset~\cite{handa2014}, each for x64-based systems and ARM-based systems. The results presented in Fig.~\ref{fig:slambench} show a comparison of boxplots of the total time elapsed for the processing of one frame. The processing of one frame includes pre-processing, tracking, integration and raycast as well as the acquisition and rendering of a single frame of the dataset. The dataset consists of \textit{881} frames and the experiment was repeated \textit{10} times. The \textit{standard deviations} and the \textit{coefficient of variations (CVs) }are shown in Tab.~\ref{tab:msd_container_performance}. For the x64-based architecture (x64/x64-Cont) as well as for the ARM-based architecture (ARM/ARM-Cont), the experiment indicates that there is no performance loss when using containers instead of native applications, for this particular benchmark. The slightly better performance of the container can be attributed to normal fluctuations of the system utilization. Earlier studies suggest that the same observation applies to the GPU passthrough performance of LXC containers~\cite{walters2014}. Both implies that the use of virtualized robot applications has no impact on the comparability and transferability of the experiments as well as on the computational performance of the underlying hardware, which is a necessary property for the success and a further dissemination of the approach.
\begin{table}
 \setlength\extrarowheight{2.5pt}
 \footnotesize
   \caption{Measures of Statistical Dispersion of the Network Performance Experiment}
    \label{tab:msd_network_performance}
    \centering
    \begin{tabular}{|p{4.3cm}|l|l|}
    \hline
    \textit{Metric} & \textit{Std. Deviation} & \textit{CV}\\ \hline\hline
    \textit{TCP\_RR\_Cont} & 51.5990 & 0.1237  \\
    \textit{TCP\_RR} & 57.1165 & 0.1280 \\
   \textit{UDP\_RR\_Cont} & 83.4594 & 0.2029 \\
   \textit{UDP\_RR} & 78.3588 & 0.1806 \\
    \textit{TCP\_Stream\_Cont} & 12.3056 & 0.1279 \\
    \textit{TCP\_Stream} & 12.4611 & 0.1280 \\
    \textit{UDP\_Steam\_Cont} & 14.2138 & 0.1469 \\
    \textit{UDP\_Stream} & 14.5790 & 0.1425 \\ \hline
    \end{tabular}
\end{table}
\subsection{Network Performace}
The following section discusses the impact on the network performace when using virtualized networks on top of physical infrastructures. In order to assure meaningful results, several different aspects of the network performance were measured, each for the native and the virtualized case. For simulating the message exchange between robots, the request/response performance of TCP (\textit{TCP\_RR}) respectively UDP (\textit{UDP\_RR}) were measured. For simulating streaming data, the streaming performance of TCP (\textit{TCP\_Stream}) and UDP (\textit{UDP\_Stream}) were measured. To reflect the real-world use case, the experiment was performed on two hosts, which are connected wirelessly. The measurements were conducted using \textit{netperf}, which is an open-source network benchmark tool. The experiment was repeated a \textit{100} times. The statistical dispersions are shown in Tab.~\ref{tab:msd_network_performance}. The results reveal that the network performance in the native case is marginal better than the performance in the container case. For \textit{TCP\_RR}, the average deviation is \textit{5.64} percent and \textit{5.45} percent for \textit{UDP\_RR}. The average streaming performance of \textit{TCP} is \textit{1.12} percent better in the native case than the streaming performance in the container case. In the case of UDP, the streaming performance in the native case is \textit{5.76} percent better than the performance in the container case. It is important to note that, because UDP is connectionless,  the streaming performance was determined using the received throughput of the receiver. It can be concluded that the performance loss of the virtualized connection is so small that it can be neglected in practice and is thus no limiting factor for the approach.
\begin{figure}[b]
\centering
\includegraphics[width=0.43\textwidth]{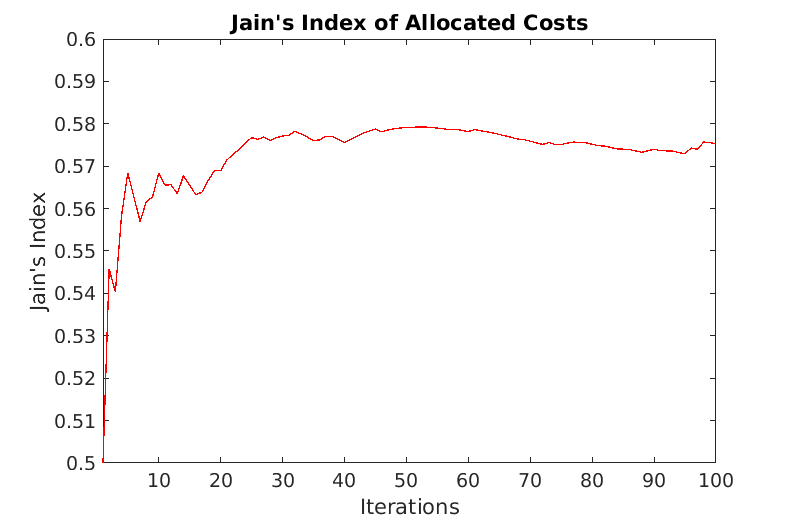}
\caption{Time Series of the Jain's Index of Allocated Costs - The x-axis represents the time in iterations and the y-axis represents the Jain's Index}
\label{fig:jain_index}
\end{figure}
\subsection{Optimality and Fairness}
\begin{figure}
\centering
\includegraphics[width=0.45\textwidth]{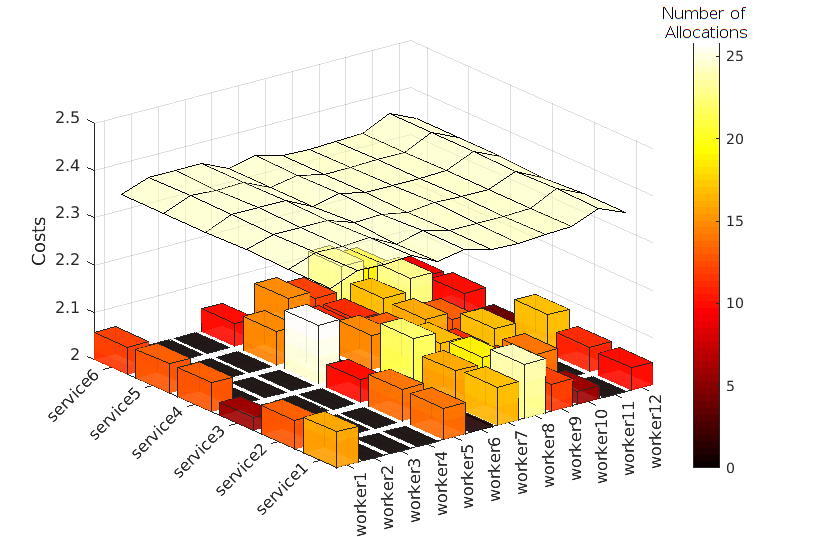}
\caption{Surface Plot of Average Cost and Bar Plot of Allocations - The surface of the plot indicates the average cost per worker, per service and the color and height of the bars indicate the number of allocations per worker, per service}
\label{fig:alloc_costs}
\end{figure}

The experiment regarding the optimality and the fairness of the approach were performed a \textit{100} times on a \textit{12}-node ARM-based cluster with a balanced workload. The optimality of the approach is shown in Fig.~\ref{fig:alloc_costs}. The figure illustrates the allocation of \textit{6} services to \textit{12} workers over \textit{100} iterations. The colorbar indicates the number of services a worker got allocated. The surface represents the average calculated costs over 100 iterations per worker, per service. The standard deviation over all costs is \textit{0.0156} and the coefficient of variation over all costs is \textit{0.0066}. A comparison of the allocations and the average costs shows that the services are allocated to the workers with the lowest costs and accordingly the lowest workload, which implies an optimal allocation of services to workers. The figure also reveals that, for this particular case, \textit{7} workers carry out the whole workload of the experiment, which seems not to be optimal with respect to the overall fairness of the approach. The \textit{Jain's Index}~\cite{jain1998} is a fairness measure that is typically used to reveal fair share of system resources. In this case, a large variance of allocated costs between the workers would result in a \textit{Jain's Index} that tends to 0. A small variance of costs and thus a balanced and fair allocation tends to a \textit{Jain's Index} of 1. The assumption that the approach not seems to be fair is supported by the results shown in Fig.~\ref{fig:jain_index}. It reveals that approx. 58\% of the workers consume 100\% of the costs. For future iterations of the system, a much better trade-off between optimality and fairness is preferable.
\subsection{Runtime}
\begin{figure}
\centering
\includegraphics[width=0.45\textwidth]{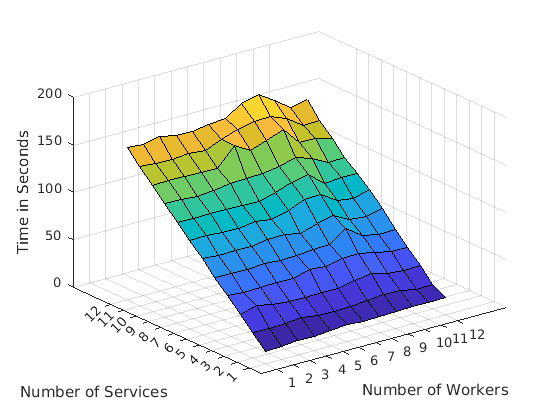}
\caption{Surface Plot of the Elapsed Time in Seconds - The figure illustrates the elapsed time in seconds from the start of the swarm to the start of the services, per service and per worker}
\label{fig:overall_eval}
\end{figure}

The following section considers the runtime of the approach. Because the time that is needed to join the swarm is very individual with regard to the environment (e.g. SSH-Access, Physical Access etc.), the comparable measure is the elapsed time from the start of the swarm to the start of the containers. The evaluation was performed for \textit{1} to \textit{12} workers and \textit{1} to \textit{12} services with the previously described cluster configuration. The result of the evaluation is shown in Fig.~\ref{fig:overall_eval}. The evaluation reveals that the time from the start of the swarm to the start of the container depends much more on the number of services that should be allocated than on the number of workers that are part of the swarm. To be precise, because of the sequential determination of costs for each service, the runtime of the approach increases linearly with the number of services. The costs for each service are calculated in a distributed and parallelized fashion, which results in a nearly constant runtime also in case of an increasing number of workers.

\section{CONCLUSIONS AND FUTURE WORK}
The present study discusses the increasing importance of reproducibility in robotics research, the approaches of the community to improve the situation and the challenges that still remains. The presented toolkit, called \textit{SwarmRob}, uses virtualization of robotics applications embedded in a novel experimental workflow to manage some of the challenges. Unlike other approaches discussed in the related work, \textit{SwarmRob} is designed to be open, flexible and extensible and is therefore easily adoptable for other researchers. The toolkit assists the researcher with the implementation of the workflow by abstracting the complexity of the used technologies in terms of orchestration, networking and initialization. The performed evaluations show that the use of containers in robotics has only a little impact on the performance of the underlying hardware, which is a critical requirement to ensure, on the one hand, the transferability of results and, on the other hand, the universal application capabilities of the toolkit. The evaluations also demonstrate the substantial time saving of the system, compared to manual experiment setups. The evaluation of the orchestration mechanism shows an optimal allocation of services to workers, which is to the detirement of fairness. Regarding this point, the measurements reveal opportunities for improvement. While currently, the initial definition of services and experiments is a cumbersome task, in future iterations, the system should be extented by a mechanism that allows to ``snapshot'' an entire robot and experiment automatically. This feature would follow up on the goal of an independent an easily accessible toolkit for reproducibility in experimental robotics research. While the system is currently under practical evaluation in a multi-robot search scenario~\cite{Sprute2017}, it should also be evaluated and evolved by the help and guidance of the community. Therefore, the entire toolkit will be publicly available at \url{https://iot-lab-minden.github.io/SwarmRob/}.


%%%%%%%%%%%%%%%%%%%%%%%%%%%%%%%%%%%%%%%%%%%%%%%%%%%%%%%%%%%%%%%%%%%%%%%%%%%%%%%%

%%%%%%%%%%%%%%%%%%%%%%%%%%%%%%%%%%%%%%%%%%%%%%%%%%%%%%%%%%%%%%%%%%%%%%%%%%%%%%%%

%%%%%%%%%%%%%%%%%%%%%%%%%%%%%%%%%%%%%%%%%%%%%%%%%%%%%%%%%%%%%%%%%%%%%%%%%%%%%%%%
%\section*{APPENDIX}


\bibliographystyle{IEEEtran}
\bibliography{IEEEabrv,lib}

\end{document}